# SYNTACTIC ANALYSIS BASED ON MORPHOLOGICAL CHARACTERISTIC FEATURES OF THE ROMANIAN LANGUAGE

Bogdan Pătru [1,2]

[1]Department of Mathematics and Informatics, Vasile Alecsandri University of Bacau, Bacau, Romania
[2]EduSoft Bacau, Romania
[1,2]bogdan@edusoft.ro

*ABSTRACT*

*This paper refers to the syntactic analysis of phrases in Romanian, as an important process of natural language processing. We will suggest a real-time solution, based on the idea of using some words or groups of words that indicate grammatical category; and some specific endings of some parts of sentence. Our idea is based on some characteristics of the Romanian language, where some prepositions, adverbs or some specific endings can provide a lot of information about the structure of a complex sentence. Such characteristics can be found in other languages, too, such as French. Using a special grammar, we developed a system (DIASEXP) that can perform a dialogue in natural language with assertive and interogative sentences about a "story" (a set of sentences describing some events from the real life).*

## 1. INTRODUCTION

Before making a semantic analysis of natural language, we should define the lexicon and syntactic rules of a formal grammar useful in generating simple sentences in the respective language (for example, English, French, or Romanian).

We shall consider a simple grammar, having some rules for the lexicon, and some rules for the grammatical categories. The rules for the lexicon will be of the type (1):

$$G \rightarrow W \qquad (1)$$

where *G* is a grammatical category (or part of speech) and *W* is an word from a certain dictionary.

The other syntactic rules will be of the type (2):

$$G_1 \rightarrow G_2 \, G_3 \qquad (2)$$

meaning that the first grammatical category ($G_1$) forms out of the concatenation of the other two ($G_2$ and $G_3$), from the right side of the arrow.

Our simple grammar is presented in Figure 1. It contains only few Romanian words and the syntax rules constitute a subset of the Romanian syntax rules:

Lexicon:

| | | | |
|---|---|---|---|
| *Det* | → | *orice* | (any) |
| *Det* | → | *fiecare* | (every) |
| *Det* | → | *o* | (a, an (fem.)) |
| *Det* | → | *un* | (a, an (masc.)) |
| *Pron* | → | *el* | (he) |
| *Pron* | → | *ea* | (she) |
| *N* | → | *bărbat* | (man) |
| *N* | → | *femeie* | (woman) |
| *N* | → | *pisică* | (cat) |
| *N* | → | *șoarece* | (mouse) |
| *V* | → | *iubește* | (loves) |
| *V* | → | *urăște* | (hates) |
| *A* | → | *frumoasă* | (beautiful) |
| *A* | → | *deșteaptă* | (smart (fem.)) |
| *A* | → | *deștept* | (smart (masc.)) |
| *C* | → | *și* | (and) |
| *C* | → | *sau* | (or) |

Syntax:

| | | |
|---|---|---|
| *S* | → | *NP VP* |
| *NP* | → | *Pron* |
| *NP* | → | *N* |
| *NP* | → | *Det N* |
| *NP* | → | *NP AP* |
| *AP* | → | *A* |
| *AP* | → | *AP CP* |
| *CA* | → | *C A* |
| *VP* | → | *V VP* |
| *VP* | → | *V NP* |

Figure 1. Example of simple grammar for Romanian language

In Figure 1, we used these notations: *S* = sentence, *NP* = noun phrase, *VP* = verb phrase, *N* = noun, *Det* = determiner (article), *AP* = adjectival phrase, *A* = adjective, *C* = conjunction, *V* = verb, *CA* = group made up of a conjunction and an adjective.

This grammar generates correct sentences in English, such as:

- Orice femeie iubește. (Every woman loves.)
- Un șoarece urăște o pisică. (A mouse hates a cat.)
- Fiecare bărbat deștept iubește o femeie frumoasă și deșteaptă. (Every smart man loves a beautiful and smart woman.)

On the other hand, this grammar rejects incorrect phrases, such as "orice iube te un b rbat" ("any loves a man"). The previously phrases contains words only from the chosen vocabulary. However, our grammar overgenerates, that is, it generates sentences that are grammatically incorrect, such as "Ea frumoas sau de teapt iube te" ("She beautiful or smart loves"), "El iube te el" ("He loves he"), and "O b rbat iube te pisic " ("An man loves cat"), even these phrases contain words from the selected lexicon.

Also, the grammar subgenerates, meaning that there are many sentences in Romanian that grammar rejects, such as "Orice femeie iube te sau ur te un b rbat". This phrase is correct in Romanian, and contains words from the given dictionary. Also, the phrase "ni te câini ur sc o pisic " ("some dogs hate a cat"), although syntactically correct in Romanian, is not accepted because it contains words that have not been entered into our vocabulary.

The syntactic analysis or the parsing of a string of words may be seen as a process of searching for a derivation tree. This may be achieved either starting from *S* and searching for a tree with the words from the given phrase as leaves (top-down parsing) or starting from the words and searching for a tree with the root *S* (bottom-up parsing). An algorithm of efficient parsing is based on dynamic programming: each time that we analyze the phrase or the string of words, we store the result so that we may not have to reanalyze it later. For example, as soon as we have discovered that a string of words is a *NP*, we may record the result in a data structure called chart. The algorithms that perform this operation are called chart-parsers. The chart-parser algorithm uses a polynomial time and a polynomial space. In (P tru & Boghian, 2010) we developed a chart-parser, based on the Cocke, Younger, and Kasami algorithm. We presented a Delphi application that analyzes the lexicon and the syntax of a sentence in Romanian. We used a Chomsky normal form (CNF) grammar (Chomsky, 1965).

## 2. USING THE DEFINITE CLAUSE GRAMMARS

In order for our grammar not to generate incorrect sentences, we should use the notions of gender, number, case etc. specifying, for example, that "femeie" and "frumoas " have the feminine gender, and the singular number. The string "El iube te ea" is incorrect, because "ea" is in nominative case, and we should use the "pe" preposition in order to obtain the accusative case. The correct phrase is "El iube te pe ea", or even"El o iube te pe ea." ("He loves her").

If we take into account the case, grammar is no longer independent from the context: it is not true that any *NP* is equal to any other *NP* irrespective of the context. Nevertheless, if we want to work with a grammar that is independent from the context, we may split the category *NP* into two, *NPN* and *NPA*, in order to represent verbal groups in the nominative (subjective), respectively accusative (objective) case. We shall also have to split the category *Pron* into two categories, *PronN* (including "El" and *PronA* (including "pe ea" ("her"), which contains the preposition "pe" in front of the pronoun "ea")  (Russel & Norvig, 2002)

Another issue concerns the agreement between the subject and main verb of the sentence (predicate). For example, if "Eu" ("I") is the subject, then "Eu iubesc" ("I love") is grammatically correct, whereas "Eu iube te" ("I loves") is not. Then we shall have to split *NPN* and *NPA* into several alternatives in order to reach the agreement. As we identify more and more distinctions, we eventually obtain an exponential number.

A more efficient solution is to improve ("augment") the existing grammar rules by using parameters for non-terminal categories. The categories *NP* and *Pron* have parameters called *gender*, *number* and *case*. The rule for the *NP* has as arguments the variables *gender*, *number* and *case*.

This formalism of improvement is called *definite clause grammar* (DCG) because each grammar rule may be interpreted as a definite clause in Horn's logic (Pereira & Warren, 1980), (Giurca, 1997).

Using adequate predicates, a CFG (context free grammar) rule as $S \rightarrow NP\ VP$ will be written in the form of the definite clause $NP(s_1) \wedge VP(s_2) => S(s_1+s_2)$, with the meaning that if $s_1$ is *NP*, $s_2$ is *VP*, then the concatenation of $s_1$ with $s_2$ is *S*. DCGs allow us to see parsing as a logical inference (Klein, 2005). The real benefit of the DCG approach is that we can improve the symbols of categories with additional arguments, for example the rule $NP(gender) \rightarrow N(gender)$ turns into the definite clause $N(gender, s_1) \Rightarrow NP(gender, s_1)$, meaning that if $s_1$ is a noun with the gender *gender*, then $s_1$ is also a noun phrase with the same gender. Generally, we may supplement a symbol of category with any number of arguments, and the arguments are parameters that constitute the subject of unification like in the common inference of Horn's clauses (Russel & Norvig, 2002).

Even with the improvements brought by DCG, incorrect sentences may still be overgenerated. In order deal with the correct verbal groups in some situations, we shall have to split the category *V* into two subcategories, one for the verbs with no object and one for the verbs with a single object, and so on. Thus, we shall have to specify which expressions (groups) may follow each verb, that is, realize a subcategorization of that verb by the list of objects. An object is a compulsory expression that follows the verb within a verbal group.

Because there are a lot of problems in dealing with the syntactic analysis or a phrase, the time of the processing the complex situations of the texts in Romanian language, describing real life situations, we decided to use some morphological characteristic features of the Romanian words, that can be useful in order to determine the parts of the sentences.

## 2. THE IDEA FOR SYNTACTIC ANALYSIS BASED ON MORPHOLOGICAL CHARACTERISTIC FEATURES OF THE LANGUAGE

As we explain in the introduction, the classic ideas of syntactic analysis uses lexicons (dictionaries), CFG or DCG grammars and chart-parsers, as that developed by us in (P tru & Boghian, 2010).

In the previous sections, we note the following:

1. Using a CFG grammar, syntactically correct sentences (in Romanian) are accepted, as well as incorrect ones.
2. The power of the analysis system (for example a chart-parser), based on such a grammar depends on the extent of the vocabulary used.
3. The power of a chart-parser can be improved using DCG grammars. This implies to extent the set of the syntactic rules, with a lot of new rules, using special variables (like gender, case, number etc.).

As concerns the first and the last issues, let us assume that the system will not be required to analyze incorrect sentences, therefore we consider the grammar satisfactory. Regarding the second problem, it could be solved by strongly enriching the vocabulary, a fact that would require the elaboration and implementing of some data structures and searching techniques as efficient as possible, which, however, will not function in real time, in some cases. Of course, the main problem would be to write as comprehensive a grammar as possible, close to the linguistic realities of Romanian morphology, taking into consideration the diversity of forms (and meanings) that a sentence can get in the Romanian language. However, the extent and complexity of grammar leads to slowing the analysis, therefore this will not be done in real-time, in all cases.

In this section we will suggest a real-time solution, based on the idea of using:

- some words or groups of words that indicate grammatical category;
- some specific endings of the inflected words that indicate some parts of sentence.

Our idea is based on some characteristics of the Romanian language, where some prepositions or some specific endings can provide us a lot of information about the structure of a complex phrase. Such characteristics can be found in other languages, too, such as French. Using our ideas, we developed a system that uses a special grammar, which we will explain.

The morphology of the Romanian language allows the developing of a special syntactic analysis that makes full use of certain characteristics of words when they are inflected (declined or conjugated) or under different hypostases.

With a view to "understanding" a Romanian sentence, to finding the constituent parts, which would favor the translation of the sentence into another language, in real time, we suggest a simple solution, based on patterns:

- there is a minimal vocabulary of key words: prefixes1, linking words, endings;
- a relatively limited grammar is realized in which the terminals are some words, either with given endings or from another given vocabulary;
- the user is asked to respect some restrictions of sentence word order (relatively a few);
- the user is assumed to be well meaning.

In Figure 2 it is presented the general scheme for such an analyzer (Pătru & Boghian, 2012).
In Figure 2, we select the concrete case of the sentence: "Copiii cei cuminți au recitat o poezie părinților, în fața școlii." ("The good children recited a poem to their parents, in front of the school").

In stage 2 predicate nouns and adjectives are identified (introduced by pronouns or prepositions), direct or indirect objects, "inarticulate" (used without an article) (introduced by articles, prepositions and prepositional phrases respectively), adverbials; in stage 3 "articled" direct and indirect objects are identified etc.

Following the logic of processes within the analyzer, we notice that the final result (that is correct in our case, up to an additional detailing) is approximate because the system is based on the observations made by us, which are:

- In Romanian, the subject usually precedes the predicate: "copiii" ("the children") before "au recitat" ("recited");
- There are some words (or groups of words) (that we call prefixes or indicators) that introduce adverbials ("în fața..."( "in front of") = place adverbial; "fiindcă ..." ("because"), "din cauză că ..." = cause adverbial; "pentru a..." ("for")= purpose adverbial;
- The predicate nouns, the adjectives and the objects (direct and indirect) can be introduced by indicating prefixes ("lui..." (Ion, Gheorghe etc.) ("to...") = indirect object (or predicate noun),

---

[1] by prefixes we understand words or groups of words, having the role to indicate the "sense" of the next words from the phrase.

"o..." ("fat " ("girl"), "pisic " ("cat") etc.) ("a…") = direct object, if these have not been already identified as subject. Thus, it is to assume that we will say: "un b iat cite te o carte" ("a boy is reading a book") and not "o carte (e ceea ce) cite te un b iat" or "o carte este citit de un b iat" ("a book (is that which) a boy is reading" or "a book is being read by a boy"), therefore the subject precedes the object);

- Some endings offer sufficient information about the nature of the respective words (in the word 'p rin ilor' the ending '-ilor' indicates an indirect object2, just like '-ei' from 'fetei', 'mamei' indicates the same type of object, the '-ul' from 'b iatul', 'creionul' indicate however a direct object ("articled").

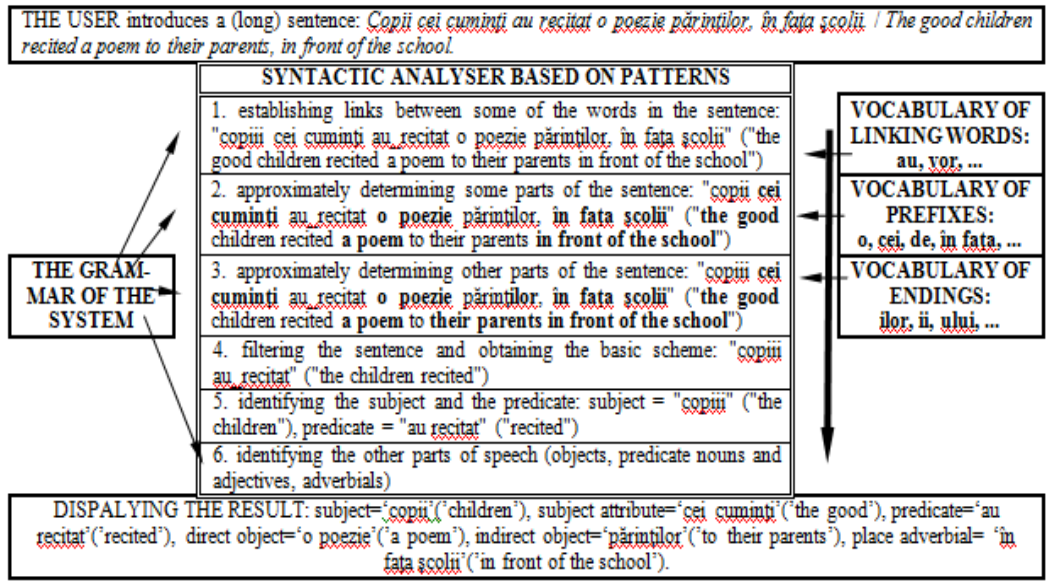

Figure 2. Stages of the analysis

We thus notice that our sentence, recognized by an analyzer of the type presented above is (structurally) complex enough as compared to the famous "orice b rbat iube te o femeie" ("any man loves a woman"), given as an example for classic chart-parsers.

It should also be noted that the sentence "I o i pe M, deoarece M e f" ("J l M, because M i b") could be recognized by the analyzer on the basis of the pattern:

- <subiect> <predicat> pe <complement direct>, deoarece <circumstan ial de cauz >.
- (<subject> <predicate>[on] <direct object>because <cause adverbial>)

Thus, the above "sentence", although meaningless in Romanian, would enter the same category as: "Ion o iube te pe Maria, deoarece Maria este frumoas " ("John loves Mary because Mary is beautiful"), a category represented by the pattern mentioned above.

---

[2] Of course, "p rin ilor" could be an indirect object as well as a predicative noun, like in the sentence "the good children recited the poems of their parents in front of the school"/ "copiii cei cumin i au recitat poeziile p rin ilor, în fa a  colii", therefore frequently not even grammar can solve the system's ambiguities.

## 3. CREATING AND CONSULTING A DATABASE

Of course, by introducing more such sentences (phrases), the user can create a table in a database with the structure of a sentence; therefore each entry would contain the fields: subject, predicate, predicative noun, direct object, indirect object, place adverbial etc. The detailed structure of the table is presented in section 5, when we will discuss about the DIASEXP shystem we developed. Consulting such a database would be made through Romanian interrogative sentences (phrases), for example:

- Cine <predicat> <complement direct> ? ("Cine cite te cartea ?") (Who <predicate><direct object>? ("Who is reading the book?"))

in order to find out the <subject>, using a search engine based on pattern-matching (matching patterns), in which the search clues would be the predicate ('is reading'), and also the direct object ('book').

Although the results of the analysis (and by this the answers to the questions also) have a high degree of precision that depends on respecting the word order restrictions imposed by the vocabulary taken into consideration, such a system that we have realised and that works in real time can be successfully used. (Moreover, the system realised by us may enrich, through learning, its vocabulary so that, by using only the 300 initial words and endings it may cover a wide range of situations).

## 4. A GENERATIVE GRAMMAR MODEL FOR SYNTACTIC ANALYSIS

We present below (Figure 3) the grammar used by our system in the syntactic analysis[3]:

| | | | |
|---|---|---|---|
| 1. | *Sentence* | → | *Subject Predicate* / *Subject Predicate Other_part_of_sen* |
| 2. | *Other_part_of_sent* | → | *Part_of_sentence* / *Part_of_sentence Other_parts_of_sent* |
| 3. | *Part_of_sentence* | → | *Adverbial* / *Object* |
| 4. | *Subject* | → | *Simple_subject* / *Simple_subject Attrib_sub* |
| 5. | *Object* | → | *Direct_object* / *Indirect_object* |
| 6. | *Adverbial* | → | *Where* / *When* / *How* / *Goal* / *Why* |
| 7. | *Direct_object* | → | *Dir_obj* / *Dir_obj Attrib_do* |
| 8. | *Indirect_object* | → | *Indir_obj* / *Indir_obj Attrib_io* |
| 9. | *Atrib_sub* | → | *Attribute* |
| 10. | *Atrrib_do* | → | *Atrribute* |
| 11. | *Atrrib_io* | → | *Atrribute* |
| 12. | *Atrribute* | → | *Adjective* / *Possesion_word* / *Pref_attrib Words T* / *Pref_attrib Word* / *Word + T_pos* / *Word + 'ând' Words T* |
| 13. | *Dir_obj* | → | *Word + T_do* / *Pref_do Word* |
| 14. | *Indir_obj* | → | *Word + T_io* / *Pref_ci Word* |
| 15. | *Where* | → | *Adv_where* / *Pref_where Words T* |
| 16. | *When* | → | *Adv_when* / *Pref_when Words T* |
| 17. | *How* | → | *Adv_how* / *Pref_how Words T* |
| 18. | *Goal* | → | *Pref_goal Words T* |
| 19. | *Why* | → | *Pref_why Words T* |
| 20. | *Pref_attrib* | → | 'al' │ 'a' │ 'ai' │ 'ale' │ 'cu' │ 'de' │ 'din' │ 'cel' │ 'cea' │ 'cei' │ 'cele' │ 'ce' │ 'care' ... |
| 21. | *Pref_do* | → | 'pe' |
| 22. | *Pref_io* | → | 'lui' │ 'de' │ 'despre' │ 'cu' │ 'cui' │ ... |
| 23. | *Pref_where* | → | 'la' │ 'în' │ 'din' │ 'de la' │ 'lâng ' │ 'în spatele' │ 'în josul' |

---
[3] the symbols from the grammar are written using Romanian

|  |  |  |
|---|---|---|
|  |  | 'spre' │ 'unde' │ 'aproape de' │ 'c tre' |
| 24. *Pref_when* | → | 'când' │ 'peste' │ 'pe' │ 'înainte de' │ 'dup ' │ ... |
| 25. *Pref_how* | → | 'altfel decât' │ 'astfel ca' │ 'în felul ' │ 'cum' │ ' a a ca' │ 'în modul' │ ... |
| 26. *Pref_goal* | → | 'pentru' │ 'pentru ca s ' │ 'în vederea' │ 'cu scopul' │ 'pentru ca' │ ... |
| 27. *Pref_why* | → | 'c ci' │ 'pentru c ' │ 'deoarece' │ 'fiindc ' │ 'întrucât' │ ... |
| 28. *Adjective* | → | 'mare' │ 'mic' │ 'bun' │ 'frumos' │ 'înalt' │ ... |
| 29. *Posession_word* | → | 'meu' │ 't u' │ 'lor' │ 'tuturor' │ 'nim nui' │ ... |
| 30. *Adv_where* | → | 'aici' │ 'acolo' │ 'dincolo' │ 'oriunde' │ ... |
| 31. *Adv_when* | → | 'acum' │ 'atunci' │ 'vara' │ 'iarna' │ 'luni' │ 'totdeauna' │ 'seara' │ ... |
| 32. *Adv_how* | → | 'a a' │ 'bine' │ 'frumos' │ 'oricum' │ 'greu' │ ... |
| 33. *T_pos* | → | 'lui' │ 'ei' │ 'ilor' │ 'elor' |
| 34. *T_do* | → | 'a' │ 'ul' │ 'ele' │ 'ile' |
| 35. *T_io* | → | 'ei' │ 'ului' │ 'elor' │ 'ilor' |
| 36. *T* | → | ',' │ '.' |
| 37. *Simple_subject* | → | Words |
| 38. *Predicate* | → | Words │ Forms_of_to_be |
| 39. *Forms_of_to_be* | → | 'este' │ 'e' │ 'sunt' │ 'e ti'… |
| 40. *Direct_obj* | → | Words |
| 41. *Indir_obj* | → | Words |
| 42. *Adjective* | → | 'mare' │ 'mic' │ 'frumos' │ 'ro u' │ … |
| 43. *Adv_where* | → | 'aici' │ 'acolo' │ 'sus' │ 'jos' │ … |
| 44. *Adv_when* | → | 'dimineața' │ 'miercuri' │ 'iarna' │ 'atunci' │ … |
| 45. *Adv_how* | → | 'a a' │ 'bine' │ 'repede' │ 'frumos' │ … |
| 46. *Words* | → | Word │ Words Words │ Words+',' Words |
| 47. *Word* | → | Character │ Word+Character |
| 48. *Character* | → | 'A' │ 'B' │ ... │ 'Z' │ 'a' │ ... │ 'z' │ '0' │ '1' │ ... │ '9' │ '-' │ ... |

Figure 3. A grammar for Romanian, based on morphological characteristics

It is to be noted that the adverbs used are the "general' ones, and the adjectives are those most frequently used in common speech. Also, please note that by "+" we noted the concatenation of two words: 'fete' + 'lor' = 'fetelor'. This concatenation can be influenced, in some cases, by a phonemic alternance, like in "fat "+"ei"="fetei", where we have the phonemic alternance a→e (see (P tru , 2010) for details).

## 4. DIASEXP

Using the grammar from the previous section, we developed the DIASEXP system. DIASEXP have a simple text interface, where the user can introduce different phrases describing some knowledge about some real life events. This collection of phrases is recorded as a "story". Each phrase of the story is analyzed by the system, which will automatically detect the parts of the sentence and will add these into a table with the following fields:

1. *Subject* – this will represent the simple subject of the sentence (a noun or a pronoun) (see rules 4 and 37 in Figure 3);
2. *Attrib_sub* – this will be the attribute of the subject (it can be, for example an adjective, see rules 9 and 12 in the same figure);
3. *Predicate* – the predicate of the sentence will represent the main action of the assertive sentence; the predicate can be represented by a normal verb or the *copulative verb* "a fi" ("to be"), which will be folowed by a *predicative noun* (*nume predicativ*) - see rules 38 and 39;
4. *Dir_obj* – the direct object or the predicative noun (see rules 7, 13, 21, and 34);
5. *Attribute_do* – this will be the attribute of the direct object (see rules 7, 10, 12, and 28)
6. *Indir_obj* – the indirect object (see rules 8, 14, 22, and 35)

7. *Attribute_io* – this will be the attribute of the indirect object (the attributes will be represented by adejectives or by some prepositional phrases) – see rules 8,11,12, and 28;
8. *Where* – this will be the place adverbial; see rules 3, 6, 15, 23, and 30;
9. *When* – this will be the time adverbial; see rules 3, 6, 16, 24, and 31;
10. *How* – this will be the manner adverbial; see rules 3, 6, 17, 25, and 32;
11. *Goal* – this will be the goal adverbial; it will be the answer at the question "Pentru ce…" ("For what…"); see rules 3, 6, 18, and 26;
12. *Why* – this will be the cause adverbial; see rules 3, 6, 19, and 27.

When DIASEXP isn't sure about which part of the sentence a word is, it will ask the user with two or three variants, and the user will indicate the correct version. The next time, in a similar situation, DIASEXP will know the answer, and it will not ask again. However, the use of commas will be very helpful, in order to avoid the ambiguities.

After entering some assertive sentences, the user can enter some interrogative sentences. When the user will ask something DIASEXP, it will answer consulting the assertive sentences it known by that moment.

Below (Table 1) you can find an example of a "story", from a dialogue between DIASEXP and a human user. The assertive sentences are prefixed by A, the interogative sentences are prefixed by I, and the DIASEXP's answers are prefixed by R. In the right column the English translation is presented.

Table 1. An example of a dialogue with DIASEXP (A=assertions, I=interogations, R=answers)

| | |
|---|---|
| A: Elena este frumoasa, deoarece are ochi frumosi. | Helen is beautiful, because she has beautiful eyes. |
| A: Elena este frumoas , deoarece are p r lung. | Helen is beautiful, because she has long hair. |
| A: Elena este frumoas , c ci e supl . | Helen is beautiful, as she is slim. |
| A: Elena este placut , întrucât e sociabil . | Helen is enjoyable, as she is sociable. |
| A: Elena e placut , deoarece e harnic . | Helen is enjoyable, because she is hard-working. |
| A: Elena este sociabil  mereu. | Helen is always sociable. |
| A: Elena e prietenoas  cu oamenii inteligenți. | Helen is amiable with intelligent people. |
| A: Elena este prietena lui Adrian. | Helen is Adrian's girlfriend. |
| A: Adrian o iubeste pe Elena, c ci e frumoas  i harnic . | Adrian loves Helen, because she is beautiful and hard-working. |
| A: Adrian nu iube te alt  fat . | Adrian doesn't love another girl. |
| A: Adrian va d rui Elenei o floare mâine, deoarece o iube te. | Adrian will give Helen a flower tomorrow because he loves her. |
| A: Elena va fi bucuroas  de floare. | Helen will be happy about the flower. |
| A: P rinții Elenei vorbesc despre Adrian. | Helen's parents talk about Adrian. |
| A: P rinții Elenei îl vor invita pe Adrian la cin , politicos. | Helen's parents will invite politely Adrian to dinner. |
| A: P rinții Elenei îl plac pe Adrian, deoarece este b iat bun. | Helen's parents like Adrian, because hs is a good boy. |
| A: Adrian este b iat bun, deoarece este student. | Adrian is a good guy, because he is a student. |
| A: Adrian, cel bun, este student la facultate, fiindc  a înv țat mult. | Adrian, the good guy, is student in college, because he learned a lot. |
| A: Adrian, studentul, este pasionat de informatic . | Adrian, the student, is passionate about computer science. |
| A: Adrian este student la informatic . | Adrian is a student of computer science. |
| A: Elena, cea frumoas , este elev  silitoare la liceu. | Helen, the beautiful one, is a sedulous student in high school. |

| A: Elena va citi azi bine lecția de istorie, pentru ca să ia notă mare. | Elena read good history lesson today, so she wants high grades. |
|---|---|
| A: Elena iubește pe Adrian. | Elena loves Adrian. |
| A: Elena iubeste pe părinții ei. | A: Elena loves her parents. |
| A: Elena s-ar căsători cu Adrian, fiindcă îl iubește. | Elena would marry Adrian because he loves. |
| A: Adrian va pleca repede, astăzi la bunicii lui, pentru ca să îi vadă. | Adrian will go quickly, today, at his grandparents to see them. |
| I: Cum este Elena? | How is Helen? |
| R: Elena este frumoasă.<br>   Elena este plăcută.<br>   Elena este elevă. | Helen is beautiful.<br>Helen is enjoyable.<br>Helen is a student. |
| I: Pe cine iubește Adrian. | Whom does Adrian love? |
| R: Adrian iubește pe Elena.<br>   Adrian nu iubește altă fată. | Adrian loves Helen.<br>Adrian doesn't love another girl. |
| I: Unde va pleca Adrian? | Where will Adrian go? |
| R: Adrian va pleca la bunicii lui? | Adrian will go to his grandparents? |
| I: Când va pleca Adrian la bunicii lui? | When will Adrian go to his grandparents? |
| R: Adrian va pleca astăzi la bunicii lui | Adrian will go today to his grandparents. |
| I: De ce este plăcută Elena? | Why is enjoyable Helen? |
| R: Elena este plăcută întrucât e sociabilă. | Helen is enjoyable because she is sociable. |
| R: Elena este placută deoarece e harnică. | Helen is enjoyable because she is hard-working. |
| Î: Cui va dărui Adrian? | Whom will gives Adrian the book? |
| R: Adrian va dărui Elenei. | Adrian will give the book to Helen. |
| I: Ce va dărui Adrian Elenei? | What will give Adrian to Helen? |
| R: Adrian va dărui Elenei o floare. | Adrian will give a flower to Helen. |

The "story" represented by the assertions from Figure 4 are stored in the table of the database like you can see in Table 2. It is to be noted that the system correctly detected the parts of sentence, for every assertions. In the table, the blank spaces correspond to those parts of sentence that are not present in that phrase.

Table 2. The table of the database, with the records of the story from Table 1

| Subject | Attrib_sub | Predicate | Dir_obj | Attribute_do | Indir_obj | Attribute_io | When | Where | How | Goal | Why |
|---|---|---|---|---|---|---|---|---|---|---|---|
| Elena |  | este | frumoasă |  |  |  |  |  |  |  | deoarece are ochi frumoși |
| Elena |  | este | frumoasă |  |  |  |  |  |  |  | deoarece are păr lung |
| Elena |  | este | frumoasă |  |  |  |  |  |  |  | căci e suplă |
| Elena |  | este | plăcută |  |  |  |  |  |  |  | întrucât e sociabila |
| Elena |  | e | plăcută |  |  |  |  |  |  |  | deoarece e harnică |
| Elena |  | este | sociabil |  |  |  | mereu |  |  |  |  |
| Elena |  | este | prietenoas |  | cu oame | Inteli |  |  |  |  |  |

| | | | | nii | genți | | | | |
|---|---|---|---|---|---|---|---|---|---|
| Elena | | este | prietena | lui Adrian | | | | | |
| Adrian | | o iubește | pe Elena | | | | | | căci e frumoasă și harnică |
| Adrian | | nu iubește | fată | altă | | | | | |
| Adrian | | va dărui | o floare | | Elenei | mâine | | | deoarece o iubește |
| Elena | | va fi | bucuroasă | | de floare | | | | |
| părinții Elenei | | vorbesc | | | despre Adrian | | | | |
| părinții Elenei | Elenei | îl vor invita | pe Adrian | | | | la ei | politicos | |
| părinții Elenei | Elenei | îl plac | pe Adrian | | | | | | deoarece este băiat bun |
| Adrian | | este | băiat | bun | | | | | deoarece este student |
| Adrian | cel bun | este | student | | | | la facultate | | fiindcă a învățat mult |
| Adrian | studentul | este | pasionat | | de informatică | | | | |
| Elena | cea frumoasă | este | elev | silitoare | | | la liceu | | |
| Elena | | iubește | pe Adrian | | | | | | |
| Elena | | iubește | pe părinții | ei | | | | | |
| Elena | | s-ar casatori | | | cu Adrian | | | | fiindcă îl iubeste |
| Adrian | | va pleca | | | | astăzi | la bunicii lui | repede | pentru ca sa ii vada |

The system was developed in Pascal programming language and was tested by our team on different situations. In graph from Figure 3 you can see the results of analyzing 4 stories, after entering respectively 20, 50, 100, and 300 sentences. The average of good results is over 80%.

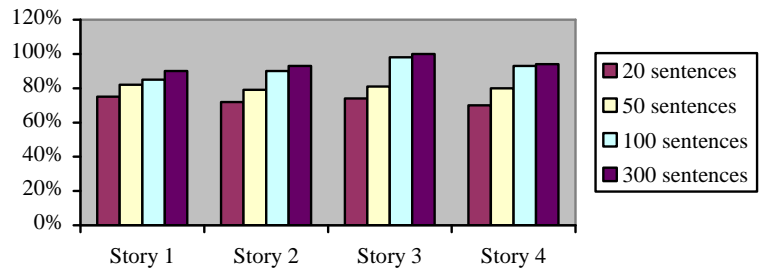

Figure 3. Results of testing DIASEXP on different stories

## 5. CONCLUSIONS

CFG and DCG are types of generative grammars used in the syntactic analysis of a phrase in natural language. Sometimes, long or complex phrases will be a problem for the classic chart-parsers. The characteristic features of the Romanian language related to the morphology of words or the way in which adverbials are formed can be successfully used in the syntactic analysis by using a system based on patterns. This system will work in real-time and will not record the whole dictionary of Romanian language. It will use a small dictionary of "prefixes" (prepositions, adverbs etc.), and some endings and linking words.

## ACKNOWLEDGEMENTS

The authors would like to thank to professors Dan Cristea, Grigor Moldovan and Ioan Andone for their advices and observations during the research activities.

**Author**

**Bogdan Pătru**

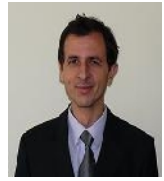

Bogdan Pătru is associate professor in computer science at Vasile Alecsandri University of Bacau, Romania, with a Ph D in computer science and a Ph D in accounting. His domains of interest/ research are natural language processing, multi-agent systems and computer science applied in social, economic and political sciences. He published papers in international journals in these domains. Also, Dr. Pătru is the author or editor of more books on programming, algorithms, artificial intelligence, interactive education, and social media.